\documentclass[a4paper]{article}

\usepackage{INTERSPEECH2020}
\usepackage{graphicx}
\usepackage{amsmath} 
\usepackage{tabularx}
\usepackage{multirow}
\usepackage{soul}
\title{Multi-modal fusion with gating using audio, lexical and disfluency features for Alzheimer's Dementia recognition from spontaneous speech}

\name{Morteza Rohanian$^1$, Julian Hough$^1$, Matthew Purver$^{1,2}$}
\address{
$^1$Cognitive Science Group\\School of Electronic Engineering and Computer Science\\
  Queen Mary University of London\\
  $^2$Department of Knowledge Technologies, Jožef Stefan Institute}
\email{ \{m.rohanian, j.hough, m.purver\}@qmul.ac.uk}

\begin{document}

\maketitle
\begin{abstract}
  
This paper is a submission to the Alzheimer's Dementia Recognition through Spontaneous Speech (ADReSS) challenge, which aims to develop methods that can assist in the automated prediction of severity of Alzheimer's Disease from speech data. We focus on acoustic and natural language features for cognitive impairment detection in spontaneous speech in the context of Alzheimer's Disease Diagnosis and the mini-mental state examination (MMSE) score prediction. We proposed a model that obtains unimodal decisions from different LSTMs, one for each modality of text and audio, and then combines them using a gating mechanism for the final prediction. We focused on sequential modelling of text and audio and investigated whether the disfluencies present in individuals’ speech relate to the extent of their cognitive impairment. Our results show that the proposed classification and regression schemes obtain very promising results on both development and test sets. This suggests Alzheimer's Disease can be detected successfully with sequence modeling of the speech data of medical sessions.
\newline

\end{abstract}


\section{Introduction}

Alzheimer’s Disease (AD) is a chronic neurodegenerative condition and the most common form of dementia. AD gradually affects the memory, language and cognitive skills and ultimately the ability to perform basic tasks in the everyday lives of patients. Early diagnosis of AD has become essential in disease management as it has not been possible to reverse the degenerative process, even with significant efforts focused on therapies \cite{67e07fa8abf2436489f5095cbe724163}.

Discrepancies in speech comprehension, speech production and memory functions are closely tied in with AD as suggested by a decrease in global vocabulary and a loss in evocative memory \cite{kempler2005neurocognitive}.
Patients with AD have difficulty performing tasks that leverage semantic information; they exhibit problems with verbal fluency and identification of objects \cite{lopez2013selection}. The semantics and pragmatics of their language appear affected throughout the entire span of the disease more than syntax \cite{bayles1982potential}. AD Patients talk more gradually with longer pauses and invest extra time seeking the right word, which contributes to disfluency of speech \cite{lopez2013selection}.

AD diagnosis demands the existence of cognitive dysfunction to be validated by neuropsychological assessments like the mini mental state examination (MMSE) performed in medical clinics \cite{folstein1975mini}. Diagnosis is typically based on the clinical analysis of patients’ history and the presence of typical neurological and neuropsychological features. It is costly and not accessible to all patients who have concerns about their memory functions.  

Recent experimental research has looked at AD's automated analysis from multimodal data as alternative, less invasive tools for diagnostics. Studying behaviours of individuals could also help detect AD earlier. There has been research on building systems which use a broad range of multimodal features to identify AD severity. A meaningful association between MMSE scores and language measures such as articulation and disfluency has been found \cite{weiner2008language}.

Much of the work to date has looked separately at the properties of the language of an individual: acoustic and lexical characteristics of speech, or syntax, fluency, and content of information. Usually these are studied within language tasks in specific domains or in conversational dialogue.
Several studies have suggested various forms of speech analysis to identify AD. Researchers found that the number of pauses, pause proportion, phonation time, phonation–to-time ratio, speach rate, articulation rate, and noise-to-harmonic ratio correlate with the severity of AD \cite{szatloczki2015speaking}. Weiner et al. \cite{weiner2016speech} developed a Linear Discriminant Analysis (LDA) classifier with a set of acoustic features such as the mean of silent segments, speech and silence durations and silence to speech ratio to distinguish subjects with AD from the control group and achieved a classification accuracy of 85.7 percent. Ambrosini et al. \cite{ambrosini2019automatic} showed an accuracy of 73 percent when using selected acoustic features (pitch, voice breaks, shimmer, speech rate, syllable duration) to detect mild cognitive impairment from a spontaneous speech task.

In terms of the features which aid AD detection, lexical features from spontaneous speech are shown to be informative. Jarrold et al. \cite{jarrold2014aided} extracted the frequency occurrence of 14 different part of speech features and combined them with acoustic features. Abel et al. \cite{abel2009connectionist} modeled patient speech errors (naming and repetition disorders) to the problem of AD diagnosis. 

There has also been work on modelling multimodal input for AD detection. Gosztolya et al. \cite{gosztolya2019identifying} examined the fusion of two SVM models with separate feature sets. The first model used a set of acoustic features, and the second model was developed using linguistic features extracted from manually annotated transcripts. Their work showed the complementary information that audio and lexical features may contain about a subject with AD. 

Among other similar tasks, using multimodal fusion to predict a cognitive state, research has been done on integrating temporal information from two or more modalities in a recurrent approaches to classify emotions or detecting different mental states, such as depression \cite{rohanian2019detecting}. One key challenge these models have is addressing the various predictive capacity of each modality and their different levels of noise. The application of a gating mechanism in various multimodal tasks has been shown to be successful in controlling the level of contribution of each modality to the eventual prediction.

This paper addresses AD classification and MMSE score regression tasks, which are part of the Alzheimer's Dementia Recognition through Spontaneous Speech (ADReSS) challenge \cite{luz2020alzheimer}. In ADReSS, participants are required to assess the AD severity of different subjects, where the target severity is based on their MMSE scores.  

We performed a binary classification of samples of speech into AD and non-AD classes and create regression models to predict MMSE scores. Using the ADReSS Challenge data which consists of speech recordings and transcripts of spoken picture descriptions, we explored various features as diagnostically relevant tools. We focused in particular on sequential modelling of sessions and whether the disfluencies and self-repairs present in individuals' speech can help predict the level of cognitive impairment.

Our approach is motivated by \cite{rohanian2019detecting} that developed the ability to learn difficult decision boundaries which other models with different methods of fusion have trouble managing, and maximise the use and combination of each modality. We employed data of individuals under controlled conditions, and modeled the sessions with audio and text features in a Long-Short Term Memory (LSTM) neural network to detect AD. Our findings indicate that AD can be detected with minimal information available on the structure of the description tasks by pure sequential modelling of a session. We also found that disfluency markers have predictive power for AD recognition.

\section{Proposed Approach}

Our approach is to model the speech of individuals giving picture descriptions as a sequence to predict whether they have AD or not, and if so, to what degree. To predict AD, we performed three sets of experiments using features from the audio and text data:

\begin{itemize}
    \item[1] LSTM models utilising unimodal audio and text features.
    \item[2] LSTM model with gating to test the effect of using multimodality. 
    \item[3] A multimodal LSTM model using acoustic and lexical information, including disfluency tagging.
\end{itemize}

The details of the three experiments are outlined below in the following sub-sections. In line with the standard assumption in deep learning, we take the approach that for a model to be genuinely data-driven, minimal feature engineering is required. The model's power is in its capacity to represent information through non-linear transforms, at varying spatial and temporal units, and from different modalities. Since we were interested in modelling temporal session changes, we used a bi-directional Long Short-Term Memory (LSTM) neural network as it has the added benefit of sequential data modelling. For each of the audio and text modalities we trained an LSTM model separately, using the audio and text features.

\subsection{Multimodal Features}

\textbf{Lexical Features from Text} A pre-trained GloVe model \cite{pennington2014glove} was used to extract the lexical feature representations from the picture description transcript and convert the utterance sequences into word vectors. We selected the hyperparameter values, which optimised the output of the model on the training set. The optimal dimension of the embedding was found to be 100.

\textbf{Audio Features} A set of 79 audio features were extracted using the COVAREP acoustic analysis framework software, a package used for automatic extraction of features from speech \cite{degottex2014covarep}. We sampled the audio features at 100Hz and used the higher-order statistics (mean, maximum, minimum, median, standard deviation, skew, and kurtosis) of COVAREP features. The features include prosodic features (fundamental frequency and voicing), voice quality features (normalized amplitude quotient, quasi open quotient, the difference in amplitude of the first two harmonics of the differentiated glottal source spectrum, maxima dispersion quotient, parabolic spectral parameter, spectral tilt/slope of wavelet responses, and shape parameter of the Liljencrants-Fant model of the glottal pulse dynamics) and spectral features (Mel cepstral coefficients 0-24, Harmonic Model and Phase Distortion mean 0-24 and deviations 0-12). Segments without audio data were set to zero. A standard zero-mean and variance normalization was applied to features. We omitted all features with no statistically significant univariate correlation with the results of training set.

\subsection{Sequence Modeling}

The potential of neural networks lies in the power to derive representations of features by non-linear input data transformations, providing greater power than traditional models. As we were interested in modelling temporal nature of speech recordings and transcripts, we used a bi-directional LSTM. For each of the audio and text modalities we trained a separate unimodal LSTM model, using different sets of features. For the input data we explored different timesteps and strides. After exploring different hyper-parameters, the model using audio data has a timestep of 20 and stride of 1 with 4 bi-directional LSTM layers with 256 hidden nodes. The model using text input has an input with a timestep of 10 and stride of 2 and has 2 LSTM layers with 16 hidden nodes. The code used in the experiments are publicly available in an online repository.\footnote{https://github.com/mortezaro/ad-recognition-from-speech}

\subsection{Multimodal Fusion with Gating}

Audio and text features can include not only discriminative and temporarily changing information about the current state of a subject, but supporting information as well. 

The model consists of two branches of the LSTM, one for each of the modalities, with their outputs combined into final feed-forward highway layers. The branches are made up of different hyperparameters and configured with respect to each modality's properties. Their outputs are concatenated and passed through N highway layers (where the best value N was determined from optimizing on heldout data). We pad the size of the training examples in the text set (which was the smaller set) to meet the audio set by mapping together instances that occurred in the same session, as the audio and text inputs for each branch of the LSTM had different timesteps and strides.

\textbf{Gating Mechanism} Data from two modalities affect the final output differently, and it is important to consider the amount of noise when aggregating them into a single representation. Since learned representation for the text can be undermined by corresponding audio representation, during multimodal fusion we need to minimise the effects of noise and overlaps. We use feed-forward highway layers \cite{srivastava2015highway}, with gating units that learn by weighing text and audio inputs at each time step to regulate information flow through network work. 

Each highway layer consists of two non-linear transformations: a Carry ($Cr$) and a Transform ($Tr$) gate which determine the degree to which the output is generated by transforming and carrying the input. Each layer uses the gates and feed-forward layer $H$ to regulate its input vector at timestep $t$, $D_t$, to generate output $y$:

\begin{equation}
y = Tr \cdot H + Cr \cdot D_t    
\end{equation}
\noindent
where $Cr$ is simply defined as $1 - Tr$, giving:
\begin{equation}
y = Tr \cdot H + (1-Tr) \cdot D_t    
\end{equation}
\noindent

The transform gate $Tr$ is defined as $\sigma(W_{Tr} D_t+ b_{Tr})$, where $W_{Tr}$ is the weight matrix and $b_{Tr}$ the bias vector for the gates. Based on the transform gates outputs, highway layers adjusts their performance from multiple-unit layers to layers that only pass through their inputs. As inspired by \cite{srivastava2015highway} and to help resolve long-term learning dependencies faster we initialise $b_{Tr}$ with a negative value (biased towards the Carry gate). We use a block of 3 stacked highway layers. The overall architecture of the LSTM with Gating model is shown in Figure 1.

\begin{figure}[!t]
\centering

\includegraphics[height=8cm]{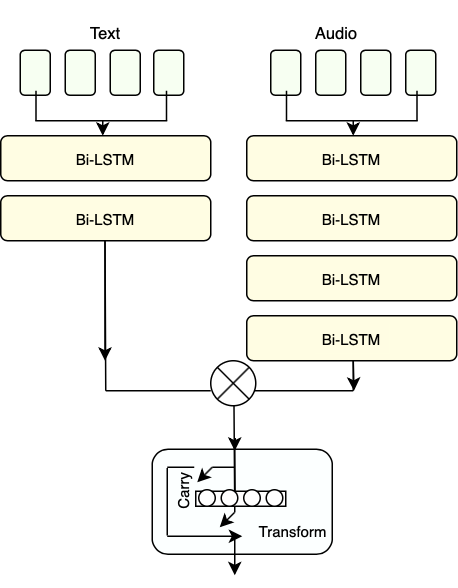}
\caption{Multimodal fusion with gating.}\label{fig:network}
\end{figure}

\subsection{Multi-modal Model with Disfluency Markers}

Disfluencies like self-repairs, pauses and fillers are widespread in everyday speech \cite{schegloff1977preference}. Disfluencies are usually seen as indicative of communication problems, caused by production or self-monitoring issues \cite{levelt1983monitoring}. Individuals with AD are likely to deal with troubles in language and cognitive skills. Patients with AD speak more slowly and with longer breaks, and invest extra time seeking the right word, which in effect contributes to disfluency \cite{lopez2013selection}. The present research explores the disfluencies present in the speech of AD patients as they contribute to severity of symptoms.

\textit{Self-repair} disfluencies are typically assumed to have a reparandum-interregnum-repair structure, in their fullest form as speech repairs \cite{shriberg1994preliminaries}. A reparandum is a speech error subsequently fixed by the speaker; the corrected expression is a repair. An interregnum word is a filler or a reference expression between the words of repair and reparandum, often a halting step as the speaker produces the repair, giving the structure as in (\ref{eg:RepairAnnotationStructure})

\vspace*{-0.3cm}
\begin{footnotesize}
\begin{equation}\label{eg:RepairAnnotationStructure}
\strut \texttt{ John }
\underbrace{\strut \texttt{ [ likes } }_\texttt{reparandum}+
\underbrace{\strut \texttt{ \{~uh~\}} }_\texttt{interregnum}
\underbrace{\strut \texttt{  loves ]} }_\texttt{repair}
\strut \texttt{ Mary}
\end{equation}
\end{footnotesize}
\vspace*{-0.2cm}

In the absence of reparandum and repair, the disfluency reduces to an isolated \textit{edit term}. A marked, lexicalized edit term such as a filled pause (``uh'' or ``um'') or more phrasal terms like ``I mean" and ``you know" can  occur. Recognizing these elements and their structure is then the task of disfluency detection.

We automatically annotated self-repairs using a deep-learning-driven model of incremental detection of disfluency developed by Hough and Schlangen \cite{HoughSchlangen15InterSpeech, hough2017joint}. It consists of deep learning sequence models that use word embeddings of incoming words, part-of-speech annotations, and other features in a left-to-right, word-by-word manner to predict disfluency tags. Here each word is either tagged as a repair onset tag (marking first word of the repair phase) edit term, or fluent word by the disfluency detector- we concatenate the disfluency tags with the word vectors to create the input for text-based LSTM.

\section{Experiments}

\subsection{Data}

The ADReSS challenge's data consists of speech recordings and transcripts of spoken picture descriptions gathered from participants via the Boston Diagnostic Aphasia Exam's Cookie Theft picture \cite{luz2020alzheimer}. The training set includes 108 subjects, and the state of the subjects is assessed on the basis of the MMSE score. MMSE is a commonly used cognitive function test for older people. It involves orientation, memory, language, and visual-spatial skills tests. Scores of 25-30 out of 30 are considered as normal, 21-24 as mild, 10-20 as moderate and $<$10 as severe impairment. 

The total number of speech segments each participant had generated was 24.86 on average. The annotations for the test set were not included in the public release of the ADReSS Challenge, so all models were tested on both the development and test set. The data is pre-processed acoustically and is balanced in terms of age and gender.

\subsection{Implementation and Metrics}

We set up our model to learn the most useful information from modalities for predicting AD. All experiments are carried out without being conditioned on the identity of the speaker. The sizes of layers and the learning rates are calculated by grid search on validation test. The LSTM models were trained using ADAM \cite{kingma2014adam} with a learning rate of 0.0001. For the loss function we used Binary Cross-Entropy to model binary outcomes, and Mean Square Error (MSE) to model regression outcomes. For binary classification of AD and non-AD, we report accuracy, precision, recall, and F1 scores and for the MMSE prediction task we report the Root Mean Square Error (RMSE).

\subsection{Baseline Models}

We compare the performance of our models to the ADReSS Challenge baseline \cite{luz2020alzheimer} with an ensemble of audio features which was provided with the dataset. The baseline classification experiments were different methods of linear discriminant analysis (LDA), decision trees (DT), and support vector machines (SVM). The baseline regression experiments were different methods of DT, gaussian process regression (GPR), and SVM.

\begin{table}[!ht]
\caption{Result of the AD classification and regression experiments with our models in cross validation}
\label{tab:results-cv}
\resizebox{\columnwidth}{!}{%
\begin{tabular}{llll}
\hline
\multicolumn{1}{|l|}{Models} & \multicolumn{1}{l|}{Features} & \multicolumn{1}{l|}{Accuracy} & \multicolumn{1}{l|}{RMSE} \\ \hline
LSTM                         & Acoustic                      & 0.64                          & 6.01                      \\
LSTM                         & Lexical                       & 0.69                          & 5.42                      \\
LSTM                         & Lexical+ Dis                  & 0.73                          & 5.08                      \\
LSTM with Gating             & Acoustic + Lexical            & 0.76                          & 5.01                      \\
LSTM with Gating             & Acoustic + Lexical + Dis       & \textbf{0.77}                 & \textbf{4.98}            
\end{tabular}}
\end{table}

\begin{table}[!ht]
\caption{Result of the AD classification and regression experiments with our models against baseline models on test set}
\label{tab:results-test}
\resizebox{\columnwidth}{!}{%
\begin{tabular}{llll}
\hline
\multicolumn{1}{|l|}{Models} & \multicolumn{1}{l|}{Features} & \multicolumn{1}{l|}{Accuracy} & \multicolumn{1}{l|}{RMSE} \\ \hline
\multicolumn{4}{l}{Baseline (\cite{luz2020alzheimer})}                                                                                             \\
LDA                          & Acoustic                      & 0.625                         & -                         \\
DT                           & Acoustic                      & 0.625                         & 6.14                      \\
SVM                          & Acoustic                      & 0.563                         & 6.12                      \\
GPR                          & Acoustic                      & -                             & 6.33                      \\
\hline
\multicolumn{4}{l}{Our Models}                                                                                          \\
LSTM                         & Acoustic                      & 0.666                         & 5.93                      \\
LSTM                         & Lexical                       & 0.708                         & 5.45                      \\
LSTM                         & Lexical + Dis                  & 0.729                         & 4.88                      \\
LSTM with Gating             & Acoustic + Lexical            & 0.771                         & 4.57                      \\
LSTM with Gating             & Acoustic + Lexical + Dis       & \textbf{0.792}                & \textbf{4.54}            
\end{tabular}}
\end{table}

\section{Results}

In Table~\ref{tab:results-cv}, we present our proposed model's performance in a cross-validation setting and in Table~\ref{tab:results-test} against that of baselines models on AD detection on the provided test set. For AD detection, our proposed LSTM model with gating and disfluency features achieves an accuracy of \textbf{0.792} and RMSE of \textbf{4.54}, outperforming all the baselines. The overall findings confirm our assumption that a model with a gating structure can more efficiently minimise the errors and noise of the individual modalities.

\textbf{Effect of disfluency features} We found that disfluency tags help as features in AD detection. Adding disfluency features to the lexical features lead to improvement in both unimodal (ACC 0.70 vs.\ 0.72; RMSE 5.45 vs.\ 4.88) and multimodal models (ACC 0.77 vs.\ 0.79; RMSE 4.57 vs.\ 4.54).

\textbf{Effect of multimodality} The multimodal LSTM with gating model outperforms the single modality AD detection models in both the classification and regression tasks. A performance increase is obtained by combining textual and audio modalities with gating over single modality models (ACC 0.72 vs.\ 0.79; RMSE 4.88 vs.\ 4.54). Adding audio features improves performance despite having different steps and timesteps inputs for each LSTM branch. In terms of our competitor baselines (without the information from the manual transcripts), multimodal classifiers performed better than all the baseline models, indicating the potential benefits of multimodal fusion in AD detection. We found that while the baseline audio-based models have some discriminative capacity, sequence modelling is more accurate (ACC scores 0.67 vs.\ 0.63) and has lower (better) RMSE (5.93 vs.\ 6.12) for predicting AD.

For AD classification, the text features alone are more informative than the audio features, as using only the text modality gives a better AD prediction than utilizing unimodal audio modality sequentially (Acc scores 0.73 vs.\ 0.67; RMSE 4.88 vs.\ 5.93).

We can see that all models provide more accurate results on the test set than in cross validation. LSTM with gating models accuracy improved more than other models on the test set (RMSE 4.54 and 4.57 vs.\ 4.98 and 5.01). 

\textbf{Error analysis} The results in Table~\ref{tab:results-confusion} show that the LSTM model with gating and disfluency features obtains the highest precision and recall for both AD and non-AD classes. The model achieves F1 scores of 0.7826 for AD and 0.8000 for non-AD. The addition of gating particularly improves the recall of AD class: the LSTM model with lexical and disfluency features without gating has a recall 0.6667 for the AD class compared to the 0.7500 achieved with gating, while its 0.7910 recall for the non-AD class is not as far beneath the 0.8333 achieved by the full gating model. Depending on the application the model is used for, false negatives or false positives for AD detection will be more or less desirable, but as it stands our full gating model considerably reduces the false negatives of diagnosis whilst still marginally reducing the false positives.

\begin{table}[!ht]
\caption{Results of AD classification task on test set}
\label{tab:results-confusion}
\resizebox{\columnwidth}{!}{%
\begin{tabular}{llllll}
\hline
\multicolumn{1}{|l|}{Models}                                                                          & \multicolumn{1}{l|}{Class} & \multicolumn{1}{l|}{Precision} & \multicolumn{1}{l|}{Recall} & \multicolumn{1}{l|}{F1 Score} & \multicolumn{1}{l|}{Accuracy} \\ \hline
\multirow{2}{*}{\begin{tabular}[c]{@{}l@{}}LSTM\\ (Lexical+ Dis)\end{tabular}}                        & AD                         & 0.7619                         & 0.6667                      & 0.7111                        & \multirow{2}{*}{0.7292}       \\
                                                                                                      & non-AD                     & 0.7037                         & 0.7910                       & 0.7451                        &                               \\ \hline
\multirow{2}{*}{\begin{tabular}[c]{@{}l@{}}LSTM with Gating\\ (Acoustic + Lexical)\end{tabular}}      & AD                         & 0.7826                         & 0.7500                      & 0.7660                        & \multirow{2}{*}{0.7708}       \\
                                                                                                      & non-AD                     & 0.7600                         & 0.7917                      & 0.7755                        &                               \\ \hline
\multirow{2}{*}{\begin{tabular}[c]{@{}l@{}}LSTM with Gating\\ (Acoustic + Lexical+ Dis)\end{tabular}} & AD                         & 0.8182                         & 0.7500                      & 0.7826                        & \multirow{2}{*}{0.7917}       \\
                                                                                                      & non-AD                     & 0.7692                         & 0.8333                      & 0.8000                        &                              
\end{tabular}}
\end{table}
\section{Conclusions}

We have presented a deep multi-modal fusion model that learns the AD indicators from audio and text modalities as well as disfluency features. We trained and tested the model on audio and transcript data from individuals doing a description task under controlled conditions, and modeled the sessions with an LSTM and feed-forward highway layers as gating mechanism for AD detection. Our findings indicate that AD can be identified by pure sequential modelling of a session, with limited information available on the structure of the description tasks. We also found that markers of disfluency hold predictive power for identification of AD.

In future work we intend to study a series of language markers associated with AD severity, as well as interactions between them. In particular, we want to undertake a more principled approach to lexical markers, disfluency markers in terms of a study of self-repair and structural markers with a look at grammatical fluency. Furthermore, we want to find acoustic features that contribute more to the prediction of AD and have higher correlation with linguistic information.

\section{Acknowledgments}

Purver is partially supported by the EPSRC under grant EP/S033564/1, and by the European Union's Horizon 2020 programme under grant agreements 769661 (SAAM, Supporting Active Ageing through Multimodal coaching) and 825153 (EMBEDDIA, Cross-Lingual Embeddings for Less-Represented Languages in European News Media). The results of this publication reflect only the authors' views and the Commission is not responsible for any use that may be made of the information it contains.

\bibliographystyle{IEEEtran}

\bibliography{mybib}


\end{document}